\documentclass[11pt,letterpaper]{article}
\usepackage{emnlp2017}
\usepackage{times}
\usepackage{latexsym}
\usepackage[pdftex]{graphicx}
\usepackage[tight,footnotesize]{subfigure}
\usepackage{amsmath}
\usepackage{amsfonts}
\usepackage{algorithmic}
\usepackage{algorithm}

\DeclareMathOperator*{\argmax}{arg\,max}

\emnlpfinalcopy

\def\emnlppaperid{10}

\title{SGNMT -- A Flexible NMT Decoding Platform for Quick Prototyping of New Models and Search Strategies}

\author{Felix Stahlberg$^\dag$ \and Eva Hasler$^{\ddagger}$ \and Danielle Saunders$^\dag$ \and Bill Byrne$^{\ddagger\dag}$ \\
\\
    $^\dag$Department of Engineering, University of Cambridge, UK  \\
\\
    $^\ddagger$SDL Research, Cambridge, UK}

\date{}

\begin{document}

\maketitle

\begin{abstract}
This paper introduces SGNMT, our experimental platform for machine translation research. SGNMT provides a generic interface to neural and symbolic scoring modules ({\em predictors}) with left-to-right semantic such as translation models like NMT, language models, translation lattices, $n$-best lists or other kinds of scores and constraints. Predictors can be combined with other predictors to form complex decoding tasks. SGNMT implements a number of search strategies for traversing the space spanned by the predictors which are appropriate for different predictor constellations. Adding new predictors or decoding strategies is particularly easy, making it a very efficient tool for prototyping new research ideas. SGNMT is actively being used by students in the MPhil program in Machine Learning, Speech and Language Technology at the University of Cambridge for course work and theses, as well as for most of the research work in our group.
\end{abstract}

\section{Introduction}

We are developing an open source decoding framework called SGNMT, short for Syntactically Guided Neural Machine Translation.\footnote{\url{http://ucam-smt.github.io/sgnmt/html/}} The software package supports a number of well-known frameworks, including TensorFlow\footnote{SGNMT relies on the TensorFlow fork available at \url{https://github.com/ehasler/tensorflow}}~\cite{nn-tensorflow}, OpenFST~\cite{fst-openfst}, Blocks/Theano~\cite{nn-theano,nn-blocks}, and NPLM~\cite{nlm-nplm}. The two central concepts in the SGNMT tool are {\em predictors} and {\em decoders}. Predictors are scoring modules which define scores over the target language vocabulary given the current internal predictor state, the history, the source sentence, and external side information. Scores from multiple, diverse predictors can be combined for use in decoding.

Decoders are search strategies which traverse the space spanned by the predictors. SGNMT provides implementations of common search tree traversal algorithms like beam search. Since decoders differ in runtime complexity and the kind of search errors they make, different decoders are appropriate for different predictor constellations.

The strict separation of scoring module and search strategy and the decoupling of scoring modules from each other makes SGNMT a very flexible decoding tool for neural and symbolic models which is applicable not only to machine translation. SGNMT is based on the OpenFST-based Cambridge SMT system~\cite{hiero-ucam-pushdown}.  Although the system is less than a year old,  we have found it to be very flexible and easy for new researchers to adopt.    Our group has already integrated SGNMT into most of its research work.

\begin{table*}
\small
\centering
\begin{tabular}{|l|p{3cm}|p{3cm}|p{3cm}|p{3cm}|}
\hline
Predictor & Predictor state & \texttt{initialize($\cdot$)} & \texttt{predict\_next()} & \texttt{consume(token)} \\
\hline
NMT & State vector in the GRU or LSTM layer of the decoder network and current context vector. & Run encoder network to compute annotations. & Forward pass through the decoder to compute the posterior given the current decoder GRU or LSTM state and the context vector. & Feed back \texttt{token} to the NMT network and update the decoder state and the context vector. \\
\hline
FST & ID of the current node in the FST. & Load FST from the file system, set the predictor state to the FST start node. & Explore all outgoing edges of the current node and use arc weights as scores. & Traverse the outgoing edge from the current node labelled with \texttt{token} and update the predictor state to the target node. \\
\hline
$n$-gram & Current $n$-gram history & Set the current $n$-gram history to the begin-of-sentence symbol. & Return the LM scores for the current $n$-gram history. & Add \texttt{token} to the current $n$-gram history.\\
\hline
Word count & None & Empty & Return a cost of 1 for all tokens except \texttt{</s>}. & Empty \\
\hline
UNK count & Number of consumed UNK tokens. & Set UNK counter to 0, estimate the $\lambda$ parameter of the Poisson distribution based on source sentence features. & For \texttt{</s>} use the log-probability of the current number of UNKs given $\lambda$. Use zero for all other tokens. & Increase internal counter by 1 if \texttt{token} is UNK. \\
\hline
\end{tabular}
\caption{\label{tab:predictor-implementations} Predictor operations for the NMT, FST, $n$-gram LM, and counting modules.}
\end{table*}

We also find that SGNMT is very well-suited for teaching and student research projects.   In the 2015-16 academic year, two students on the Cambridge MPhil in Machine Learning, Speech and Language Technology used SGNMT for their dissertation 
projects.\footnote{\url{http://www.mlsalt.eng.cam.ac.uk/Main/CurrentMPhils}}
The first project involved using SGNMT with OpenFST for applying sub-word models in SMT~\cite{mphil-jiameng}.   The second project developed automatic music composition by LSTMs where WFSAs were used to define the space of allowable chord progressions in `Bach' chorales~\cite{mphil-marcin}. The LSTM provides the `creativity' and the WFSA enforces constraints that the chorales must obey. This second project in particular demonstrates the versatility of the approach.     For the current, 2016-17 academic year, SGNMT is being used heavily in two courses.

\begin{table}
\small
\centering
\begin{tabular}{|l|p{5cm}|}
\hline
\bf Predictor & \bf Description \\
\hline
nmt & Attention-based neural machine translation following Bahdanau et al.\ \shortcite{nmt-bahdanau}. Supports Blocks/Theano~\cite{nn-theano,nn-blocks} and TensorFlow~\cite{nn-tensorflow}. \\
\hline
fst & Predictor for rescoring deterministic lattices~\cite{sgnmt}. \\
\hline
nfst & Predictor for rescoring non-deterministic lattices. \\
\hline
rtn & Rescoring recurrent transition networks (RTNs) as created by HiFST~\cite{hiero-ucam-pushdown} with late expansion. \\
\hline
srilm & $n$-gram Kneser-Ney language model using the SRILM~\cite{lm-heafield,lm-srilm} toolkit. \\
\hline
nplm & Neural $n$-gram language models based on NPLM~\cite{nlm-nplm}. \\
\hline
rnnlm & Integrates RNN language models with TensorFlow as described by Zaremba et al.\ \shortcite{rnnlm-zaremba}. \\
\hline
forced & Forced decoding with a single reference. \\
\hline
forcedlst & $n$-best list rescoring. \\
\hline
bow & Restricts the search space to a bag of words with or without repetition~\cite{eva-wordordering}. \\
\hline
lrhiero & Experimental implementation of left-to-right Hiero~\cite{hiero-lr} for small grammars. \\
\hline
wc & Number of words feature. \\
\hline
unkc & Applies a Poisson model for the number of UNKs in the output. \\
\hline
ngramc & Integrates external $n$-gram posteriors, e.g.\ for MBR-based NMT according Stahlberg et al.\ \shortcite{nmt-mbr}. \\
\hline
length & Target sentence length model using simple source sentence features. \\
\hline
\end{tabular}
\caption{\label{tab:predictors} Currently implemented predictors.}
\end{table}

\section{Predictors}

SGNMT consequently emphasizes flexibility and extensibility by providing a common interface to a wide range of constraints or models used in MT research. The concept facilitates quick prototyping of new research ideas. Our platform aims to minimize the effort required for implementation; decoding speed is secondary as optimized code for production systems can be produced once an idea has been proven successful in the SGNMT framework. In SGNMT, scores are assigned to partial hypotheses via one or many predictors. One predictor usually has a single responsibility as it represents a single model or type of constraint. Predictors need to implement the following methods:

\begin{itemize}
\item \texttt{initialize(src\_sentence)} Initialize the predictor state using the source sentence.
\item \texttt{get\_state()} Get the internal predictor state.
\item \texttt{set\_state(state)} Set the internal predictor state.
\item \texttt{predict\_next()} Given the internal predictor state, produce the posterior over target tokens for the next position.
\item \texttt{consume(token)} Update the internal predictor state by adding \texttt{token} to the current history.
\end{itemize}

The structure of the predictor state and the implementations of these methods differ substantially between predictors. Tab.~\ref{tab:predictors} lists all predictors which are currently implemented. Tab.~\ref{tab:predictor-implementations} summarizes the semantics of this interface for three very common predictors: the neural machine translation (NMT) predictor, the (deterministic) finite state transducer (FST) predictor for lattice rescoring, and the $n$-gram predictor for applying $n$-gram language models. We also included two examples (word count and UNK count) which do not have a natural left-to-right semantic but can still be represented as predictors.

\subsection{Example Predictor Constellations}
\label{sec:predictor-examples}

SGNMT allows combining any number of predictors and even multiple instances of the same predictor type. In case of multiple predictors we combine the predictor scores in a linear model. The following list illustrates that various interesting decoding tasks can be formulated as predictor combinations.

\begin{itemize}
\item \texttt{nmt}: A single NMT predictor represents pure NMT decoding.
\item \texttt{nmt,nmt,nmt}: Using multiple NMT predictors is a natural way to represent ensemble decoding~\cite{nn-ensembles,nmt-sutskever} in our framework.
\item \texttt{fst,nmt}: NMT decoding constrained to an FST. This can be used for neural lattice rescoring~\cite{sgnmt} or other kinds of constraints, for example in the context of source side simplification in MT~\cite{eva-simplification} or chord progressions in `Bach'~\cite{mphil-marcin}. The {\em fst} predictor can also be used to restrict the output of character-based or subword-unit-based NMT to a large word-level vocabulary encoded as FSA.
\item \texttt{nmt,rnnlm,srilm,nplm}: Combining NMT with three kinds of language models: An RNNLM~\cite{rnnlm-zaremba}, a Kneser-Ney $n$-gram LM~\cite{lm-heafield,lm-srilm}, and a feedforward neural network LM~\cite{nlm-nplm}.
\item \texttt{nmt,ngramc,wc}: MBR-based NMT following Stahlberg et al.\ \shortcite{nmt-mbr} with $n$-gram posteriors extracted from an SMT lattice (\texttt{ngramc}) and a simple word penalty (\texttt{wc}).
\end{itemize}

\begin{table}
\small
\centering
\begin{tabular}{|l|p{5cm}|}
\hline
\bf Decoder & \bf Description \\
\hline
greedy & Greedy decoding. \\
\hline
beam & Beam search as described in Bahdanau et al.\ \cite{nmt-bahdanau}. \\
\hline
dfs & Depth-first search. Efficiently enumerates the complete search space, e.g.\ for exhaustive FST-based rescoring. \\
\hline
restarting & Similar to DFS but with better admissible pruning behaviour. \\
\hline
astar & A* search~\cite{ai-norvig}. The heuristic function can be defined via predictors. \\
\hline
sepbeam & Associates hypotheses in the beam with only one predictor. Efficiently approximates system-level combination. \\
\hline
syncbeam & Beam search which compares hypotheses after consuming a special synchronization symbol rather than after each iteration. \\
\hline
bucket & Multiple beam search passes with small beam size. Can have better pruning behaviour than standard beam search. \\
\hline
vanilla & Fast beam search decoder for (ensembled) NMT. This implementation is similar to the decoder in Blocks~\cite{nn-blocks} but can only be used for NMT as it bypasses the predictor framework. \\
\hline
\end{tabular}
\caption{\label{tab:decoders} Currently implemented decoders.}
\end{table}

\begin{figure*}[!t]
\centering
\includegraphics[width=1\linewidth]{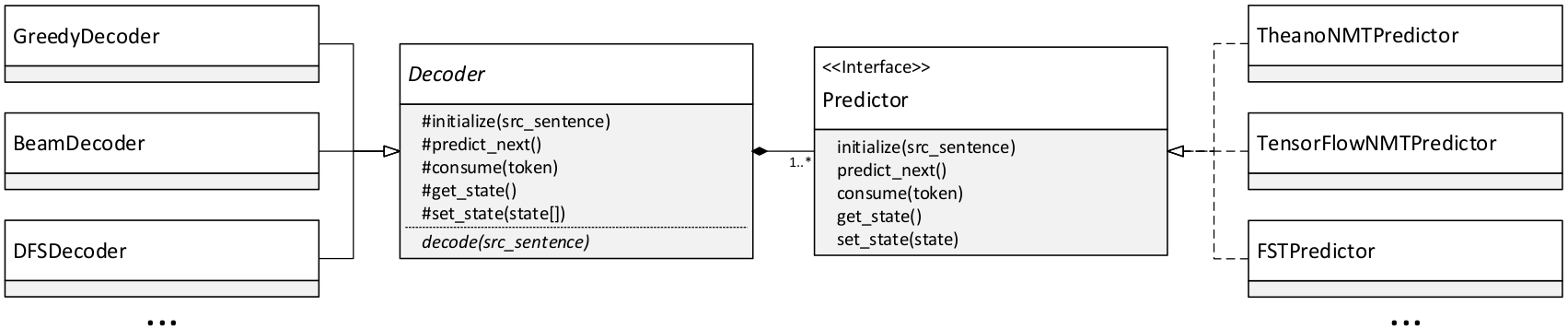}
\caption{Reduced UML class diagram.}
\label{fig:uml}
\end{figure*}

\section{Decoders}

{\em Decoders} are algorithms to search for the highest scoring hypothesis. The list of predictors determines how (partial) hypotheses are scored by implementing the methods \texttt{initialize($\cdot$)}, \texttt{get\_state()}, \texttt{set\_state($\cdot$)}, \texttt{predict\_next()}, and \texttt{consume($\cdot$)}. The {\em Decoder} class implements versions of these methods which apply to all predictors in the list. \texttt{initialize($\cdot$)} is always called prior to decoding a new sentence. Many popular search strategies can be described via the remaining methods \texttt{get\_state()}, \texttt{set\_state($\cdot$)}, \texttt{predict\_next()}, and \texttt{consume($\cdot$)}. Algs.~\ref{alg:greedy} and \ref{alg:beam} show how to define greedy and beam decoding in this way.\footnote{Formally, \texttt{predict\_next()} in Algs.~\ref{alg:greedy} and \ref{alg:beam} returns pairs of tokens and their costs.}\footnote{String concatenation is denoted with $\cdot$.}

Tab.~\ref{tab:decoders} contains a list of currently implemented decoders. The UML diagram in Fig.~\ref{fig:uml} illustrates the relation between decoders and predictors.

\paragraph{NMT batch decoding}

The flexibility of the predictor framework comes with degradation in decoding time. SGNMT provides two ways of speeding up pure NMT decoding, especially on the GPU. The {\em vanilla} decoding strategy exposes the beam search implementation in Blocks~\cite{nn-blocks} which processes all active hypotheses in the beam in parallel. We also implemented a beam decoder version which decodes multiple sentences at once (batch decoding) rather than in a sequential order. Batch decoding is potentially more efficient since larger batches can make better use of GPU parallelism. The key concepts of our batch decoder implementation are:

\begin{algorithm}[t!]
\caption{Greedy(src\_sen)}
\label{alg:greedy}
\begin{algorithmic}[1]
\STATE{\texttt{initialize}(src\_sen)}
\STATE{$h\gets\langle \texttt{<s>} \rangle$}
\REPEAT
  \STATE{$P\gets$\texttt{predict\_next()}}
  \STATE{$(t,c) \gets \argmax_{(t',c')\in P} c'$}
  \STATE{$h\gets h\cdot t$}
  \STATE{\texttt{consume}($t$)}
\UNTIL{$t=\texttt{</s>}$}
\RETURN{$h$}
\end{algorithmic}
\end{algorithm}

\begin{algorithm}[t!]
\caption{Beam($n$, src\_sen)}
\label{alg:beam}
\begin{algorithmic}[1]
\STATE{\texttt{initialize}(src\_sen)}
\STATE{$H\gets \{ (\langle\texttt{<s>}\rangle, 0.0, \texttt{get\_state()}) \}$}
\REPEAT
  \STATE{$H_{next}\gets\emptyset$}
  \FORALL{$(h,c,s)\in H$}
    \STATE{\texttt{set\_state($s$)}}
    \STATE{$P\gets$\texttt{predict\_next()}}
    \STATE{$H_{next}\gets H_{next}\cup $\\
    \hspace{2cm}$\bigcup_{(t',c')\in P}(h\cdot t', c+c',s)$}
  \ENDFOR
  \STATE{$H\gets\emptyset$}
  \FORALL{$(h,c,s)\in n\text{-best}(H_{next})$}
    \STATE{\texttt{set\_state}($s$)}
    \STATE{\texttt{consume}($h_{|h|}$)}
    \STATE{$H\gets H\cup \{(h,c,\texttt{get\_state}())\}$}
  \ENDFOR
\UNTIL{Best hypothesis in $H$ ends with \texttt{</s>}}
\RETURN{Best hypothesis in $H$}
\end{algorithmic}
\end{algorithm}

\begin{itemize}
\item We use a scheduler running on a separate CPU thread to construct large batches of computation (GPU jobs) from multiple sentences and feeding them to the {\em jobs} queue.
\item The GPU is operated by a single thread which communicates with the CPU scheduler thread via queues containing jobs. This thread is only responsible for retrieving jobs in the {\em jobs} queue, computing them, and putting them in the {\em jobs\_results} queue, minimizing the down-time of GPU computation.
\item Yet another CPU thread is responsible for processing the results computed on the GPU in the {\em job\_results} queue, e.g.\ by getting the $n$-best words from the posteriors. Processed jobs are sent back to the CPU scheduler where they are reassembled into new jobs. 
\end{itemize}

This decoder is able to translate the WMT English-French test sets {\em news-test2012} to {\em news-test2014} on a Titan X GPU with 911.6 words per second with the word-based NMT model described in Stahlberg et al.~\shortcite{sgnmt}.\footnote{Theano 0.9.0, cuDNN 5.1, Cuda 8 with CNMeM, Intel\textsuperscript{\textregistered} Core i7-6700 CPU} This decoding speed seems to be slightly faster than sequential decoding with high-performance NMT decoders like Marian-NMT~\cite{amunmt} with reported decoding speeds of 865 words per second.\footnote{Note that the comparability is rather limited since even though we use the same beam size (5) and vocabulary sizes (30k), we use (a) a slightly slower GPU (Titan X vs.\ GTX 1080), (b) a different training and test set, (c) a slightly different network architecture, and (d) words rather than subword units.} However, batch decoding with Marian-NMT is much faster reaching over 4,500 words per second.\footnote{\url{https://marian-nmt.github.io/features/}} We think that these differences are mainly due to the limited multithreading support and performance in Python especially when using external libraries as opposed to the highly optimized C++ code in Marian-NMT. We did not push for even faster decoding as speed is not a major design goal of SGNMT. Note that batch decoding bypasses the predictor framework and can only be used for pure NMT decoding.

\paragraph{Ensembling with models at multiple tokenization levels}

SGNMT allows masking predictors with alternative sets of modelling units. The conversion between the tokenization schemes of different predictors is defined with FSTs. This makes it possible to decode by combining scores from both a subword-unit (BPE) based NMT~\cite{nmt-bpe} and a word-based NMT model with character-based NMT, masking the BPE-based and word-based NMT predictors with FSTs which transduce character sequences to BPE or word sequences. Masking is transparent to the decoding strategy as predictors are replaced by a special wrapper ({\em fsttok}) that uses the masking FST to translate \texttt{predict\_next()} and \texttt{consume()} calls to (a series of) predictor calls with alternative tokens. The {\em syncbeam} variation of beam search compares competing hypotheses only after consuming a special word boundary symbol rather than after each token. This allows combining scores at the word level even when using models with multiple levels of tokenization. Joint decoding with different tokenization schemes has the potential of combining the benefits of the different schemes: character- and BPE-based models are able to address rare words, but word-based NMT can model long-range dependencies more efficiently.

\paragraph{System-level combination}

We showed in Sec.~\ref{sec:predictor-examples} how to formulate NMT ensembling as a set of NMT predictors. Ensembling averages the individual model scores in each decoding step. Alternatively, system-level combination decodes the entire sentence with each model separately, and selects the best scoring complete hypothesis over all models. In our experiments, system-level combination is not as effective as ensembling but still leads to moderate gains for pure NMT. However, a trivial implementation which selects the best translation in a postprocessing step after separate decoding runs is slow. The {\em sepbeam} decoding strategy reduces the runtime of system-level combination to the single system level. The strategy applies only one predictor rather than a linear combination of all predictors to expand a hypothesis. The single predictor is linked by the parent hypothesis. The initial stack in {\em sepbeam} contains hypotheses for each predictor (i.e.\ system) rather than only one as in normal beam search. We report a moderate gain of 0.5 BLEU over a single system on the Japanese-English ASPEC test set~\cite{data-aspec} by combining three BPE-based NMT models from Stahlberg et al.~\shortcite{nmt-mbr} using the {\em sepbeam} decoder.

\paragraph{Iterative beam search}

Normal beam search is difficult to use in a time-constrained setting since the runtime depends on the {\em target} sentence length which is a priori not known, and it is therefore hard to choose the right beam size beforehand. The {\em bucket} search algorithm sidesteps the problem of setting the beam size by repeatedly performing small beam search passes until a fixed computational budget is exhausted. Bucket search produces an initial hypothesis very quickly, and keeps the partial hypotheses for each length in buckets. Subsequent beam search passes refine the initial hypothesis by iteratively updating these buckets. Our initial experiments suggest that bucket search often performs on a similar level as standard beam search with the benefit of being able to support hard time constraints. Unlike beam search, bucket search lends itself to risk-free (i.e.\ admissible) pruning since all partial hypotheses worse than the current best complete hypothesis can be discarded.

\section{Conclusion}

This paper presented our SGNMT platform for prototyping new approaches to MT which involve both neural and symbolic models. SGNMT supports a number of different models and constraints via a common interface ({\em predictors}), and various search strategies ({\em decoders}). Furthermore, SGNMT focuses on minimizing the implementation effort for adding new predictors and decoders by decoupling scoring modules from each other and from the search algorithm. SGNMT is actively being used for teaching and research and we welcome contributions to its development, for example by implementing new predictors for using models trained with other frameworks and tools.

\section*{Acknowledgments}

This work  was  supported  by the U.K.\ Engineering and Physical Sciences Research Council (EPSRC grant EP/L027623/1).

\bibliography{refs}

\begin{thebibliography}{23}
\expandafter\ifx\csname natexlab\endcsname\relax\def\natexlab#1{#1}\fi

\bibitem[{Abadi et~al.(2016)Abadi, Agarwal, Barham, Brevdo, Chen, Citro,
  Corrado, Davis, Dean, Devin et~al.}]{nn-tensorflow}
Mart{\i}n Abadi, Ashish Agarwal, Paul Barham, Eugene Brevdo, Zhifeng Chen,
  Craig Citro, Greg~S Corrado, Andy Davis, Jeffrey Dean, Matthieu Devin, et~al.
  2016.
\newblock Tensorflow: {Large-scale} machine learning on heterogeneous
  distributed systems.
\newblock \emph{arXiv preprint arXiv:1603.04467}.

\bibitem[{Allauzen et~al.(2014)Allauzen, Byrne, de~Gispert, Iglesias, and
  Riley}]{hiero-ucam-pushdown}
Cyril Allauzen, Bill Byrne, Adri{\`a} de~Gispert, Gonzalo Iglesias, and Michael
  Riley. 2014.
\newblock Pushdown automata in statistical machine translation.
\newblock \emph{Computational Linguistics}, 40(3):687--723.

\bibitem[{Allauzen et~al.(2007)Allauzen, Riley, Schalkwyk, Skut, and
  Mohri}]{fst-openfst}
Cyril Allauzen, Michael Riley, Johan Schalkwyk, Wojciech Skut, and Mehryar
  Mohri. 2007.
\newblock {OpenFst}: A general and efficient weighted finite-state transducer
  library.
\newblock In \emph{International Conference on Implementation and Application
  of Automata}, pages 11--23. Springer.

\bibitem[{Bahdanau et~al.(2015)Bahdanau, Cho, and Bengio}]{nmt-bahdanau}
Dzmitry Bahdanau, Kyunghyun Cho, and Yoshua Bengio. 2015.
\newblock Neural machine translation by jointly learning to align and
  translate.
\newblock In \emph{ICLR}.

\bibitem[{Bastien et~al.(2012)Bastien, Lamblin, Pascanu, Bergstra, Goodfellow,
  Bergeron, Bouchard, Warde-Farley, and Bengio}]{nn-theano}
Fr{\'e}d{\'e}ric Bastien, Pascal Lamblin, Razvan Pascanu, James Bergstra, Ian
  Goodfellow, Arnaud Bergeron, Nicolas Bouchard, David Warde-Farley, and Yoshua
  Bengio. 2012.
\newblock Theano: {New} features and speed improvements.
\newblock In \emph{NIPS}.

\bibitem[{Gao(2016)}]{mphil-jiameng}
Jiameng Gao. 2016.
\newblock Variable length word encodings for neural translation models.
\newblock {MPhil} dissertation, University of Cambridge.

\bibitem[{Hansen and Salamon(1990)}]{nn-ensembles}
Lars~Kai Hansen and Peter Salamon. 1990.
\newblock Neural network ensembles.
\newblock \emph{IEEE transactions on pattern analysis and machine
  intelligence}, 12(10):993--1001.

\bibitem[{Hasler et~al.(2016)Hasler, de~Gispert, Stahlberg, Waite, and
  Byrne}]{eva-simplification}
Eva Hasler, Adri{\`a} de~Gispert, Felix Stahlberg, Aurelien Waite, and Bill
  Byrne. 2016.
\newblock Source sentence simplification for statistical machine translation.
\newblock \emph{Computer Speech \& Language}.

\bibitem[{Hasler et~al.(2017)Hasler, Stahlberg, Tomalin, de~Gispert, and
  Byrne}]{eva-wordordering}
Eva Hasler, Felix Stahlberg, Marcus Tomalin, Adri\`{a} de~Gispert, and Bill
  Byrne. 2017.
\newblock A comparison of neural models for word ordering.
\newblock In \emph{INLG}, Santiago de Compostela, Spain.

\bibitem[{Heafield et~al.(2013)Heafield, Pouzyrevsky, Clark, and
  Koehn}]{lm-heafield}
Kenneth Heafield, Ivan Pouzyrevsky, Jonathan~H. Clark, and Philipp Koehn. 2013.
\newblock \href {http://www.aclweb.org/anthology/P13-2121} {Scalable modified
  {Kneser-Ney} language model estimation}.
\newblock In \emph{ACL}, pages 690--696, Sofia, Bulgaria.

\bibitem[{Junczys-Dowmunt et~al.(2016)Junczys-Dowmunt, Dwojak, and
  Hoang}]{amunmt}
Marcin Junczys-Dowmunt, Tomasz Dwojak, and Hieu Hoang. 2016.
\newblock Is neural machine translation ready for deployment? a case study on
  30 translation directions.
\newblock \emph{arXiv preprint arXiv:1610.01108}.

\bibitem[{van Merri{\"e}nboer et~al.(2015)van Merri{\"e}nboer, Bahdanau,
  Dumoulin, Serdyuk, Warde-Farley, Chorowski, and Bengio}]{nn-blocks}
Bart van Merri{\"e}nboer, Dzmitry Bahdanau, Vincent Dumoulin, Dmitriy Serdyuk,
  David Warde-Farley, Jan Chorowski, and Yoshua Bengio. 2015.
\newblock Blocks and fuel: {Frameworks} for deep learning.
\newblock \emph{arXiv preprint arXiv:1506.00619}.

\bibitem[{Nakazawa et~al.(2016)Nakazawa, Yaguchi, Uchimoto, Utiyama, Sumita,
  Kurohashi, and Isahara}]{data-aspec}
Toshiaki Nakazawa, Manabu Yaguchi, Kiyotaka Uchimoto, Masao Utiyama, Eiichiro
  Sumita, Sadao Kurohashi, and Hitoshi Isahara. 2016.
\newblock {ASPEC}: {Asian} scientific paper excerpt corpus.
\newblock In \emph{LREC}, pages 2204--2208, Portoroz, Slovenia.

\bibitem[{Russell and Norvig(2003)}]{ai-norvig}
Stuart~J. Russell and Peter Norvig. 2003.
\newblock \emph{Artificial Intelligence: {A} Modern Approach}, 2 edition.
\newblock Pearson Education.

\bibitem[{Sennrich et~al.(2016)Sennrich, Haddow, and Birch}]{nmt-bpe}
Rico Sennrich, Barry Haddow, and Alexandra Birch. 2016.
\newblock \href {http://www.aclweb.org/anthology/P16-1162} {Neural machine
  translation of rare words with subword units}.
\newblock In \emph{ACL}, pages 1715--1725, Berlin, Germany.

\bibitem[{Siahbani et~al.(2013)Siahbani, Sankaran, and Sarkar}]{hiero-lr}
Maryam Siahbani, Baskaran Sankaran, and Anoop Sarkar. 2013.
\newblock \href {http://www.aclweb.org/anthology/D13-1110} {Efficient
  left-to-right hierarchical phrase-based translation with improved
  reordering}.
\newblock In \emph{EMNLP}, pages 1089--1099, Seattle, Washington, USA.

\bibitem[{Stahlberg et~al.(2017)Stahlberg, de~Gispert, Hasler, and
  Byrne}]{nmt-mbr}
Felix Stahlberg, Adri\`{a} de~Gispert, Eva Hasler, and Bill Byrne. 2017.
\newblock \href {http://www.aclweb.org/anthology/E17-2058} {Neural machine
  translation by minimising the {Bayes}-risk with respect to syntactic
  translation lattices}.
\newblock In \emph{EACL}, pages 362--368, Valencia, Spain.

\bibitem[{Stahlberg et~al.(2016)Stahlberg, Hasler, Waite, and Byrne}]{sgnmt}
Felix Stahlberg, Eva Hasler, Aurelien Waite, and Bill Byrne. 2016.
\newblock \href {http://anthology.aclweb.org/P16-2049} {Syntactically guided
  neural machine translation}.
\newblock In \emph{ACL}, pages 299--305, Berlin, Germany.

\bibitem[{Stolcke et~al.(2002)}]{lm-srilm}
Andreas Stolcke et~al. 2002.
\newblock {SRILM} -- an extensible language modeling toolkit.
\newblock In \emph{Interspeech}, volume 2002, page 2002.

\bibitem[{Sutskever et~al.(2014)Sutskever, Vinyals, and Le}]{nmt-sutskever}
Ilya Sutskever, Oriol Vinyals, and Quoc~V. Le. 2014.
\newblock Sequence to sequence learning with neural networks.
\newblock In \emph{NIPS}, pages 3104--3112. MIT Press.

\bibitem[{Tomczak(2016)}]{mphil-marcin}
Marcin Tomczak. 2016.
\newblock Bachbot.
\newblock {MPhil} dissertation, University of Cambridge.

\bibitem[{Vaswani et~al.(2013)Vaswani, Zhao, Fossum, and Chiang}]{nlm-nplm}
Ashish Vaswani, Yinggong Zhao, Victoria Fossum, and David Chiang. 2013.
\newblock \href {http://www.aclweb.org/anthology/D13-1140} {Decoding with
  large-scale neural language models improves translation}.
\newblock In \emph{EMNLP}, pages 1387--1392, Seattle, Washington, USA.

\bibitem[{Zaremba et~al.(2014)Zaremba, Sutskever, and Vinyals}]{rnnlm-zaremba}
Wojciech Zaremba, Ilya Sutskever, and Oriol Vinyals. 2014.
\newblock Recurrent neural network regularization.
\newblock \emph{arXiv preprint arXiv:1409.2329}.

\end{thebibliography}


\begin{thebibliography}{4}
\expandafter\ifx\csname natexlab\endcsname\relax\def\natexlab#1{#1}\fi

\bibitem[{Aho and Ullman(1972)}]{Aho:72}
Alfred~V. Aho and Jeffrey~D. Ullman. 1972.
\newblock \emph{The Theory of Parsing, Translation and Compiling}, volume~1.
\newblock Prentice-Hall, Englewood Cliffs, NJ.

\bibitem[{{American Psychological Association}(1983)}]{APA:83}
{American Psychological Association}. 1983.
\newblock \emph{Publications Manual}.
\newblock American Psychological Association, Washington, DC.

\bibitem[{Chandra et~al.(1981)Chandra, Kozen, and Stockmeyer}]{Chandra:81}
Ashok~K. Chandra, Dexter~C. Kozen, and Larry~J. Stockmeyer. 1981.
\newblock \href {https://doi.org/10.1145/322234.322243} {Alternation}.
\newblock \emph{Journal of the Association for Computing Machinery},
  28(1):114--133.

\bibitem[{Gusfield(1997)}]{Gusfield:97}
Dan Gusfield. 1997.
\newblock \emph{Algorithms on Strings, Trees and Sequences}.
\newblock Cambridge University Press, Cambridge, UK.

\end{thebibliography}
\bibliographystyle{emnlp_natbib}

\end{document}